# An approach to human iris recognition using quantitative analysis of image features and machine learning

[1]Abolfazl Zargari Khuzani[*], [2]Najmeh Mashhadi, [3]Morteza Heidari, [4]Donya Khaledyan
[1]The Department of Electrical and Computer Engineering, University of California, Santa Cruz, USA
[2]The Department of Computer Science and Engineering, University of California, Santa Cruz, USA
[3]School of Electrical & Computer Engineering, University of Oklahoma, Norman, USA
[4]Faculty of Electrical Engineering, Shahid Beheshti University, Tehran, Iran
* abzargar@ucsc.edu

*Abstract*— **The Iris pattern is a unique biological feature for each individual, making it a valuable and powerful tool for human identification. In this paper, an efficient framework for iris recognition is proposed in four steps. (1) Iris segmentation (using a relative total variation combined with Coarse Iris Localization), (2) feature extraction (using Shape&density, FFT, GLCM, GLDM, and Wavelet), (3) feature reduction (employing Kernel-PCA) and (4) classification (applying multi-layer neural network) to classify 2000 iris images of CASIA-Iris-Interval dataset obtained from 200 volunteers. The results confirm that the proposed scheme can provide a reliable prediction with an accuracy of up to 99.64%.**

## I. Introduction

The iris is a part of the eye that controls the pupil's size, regulating the amount of light that enters the eye. It is the part of the eye with coloring based on the amount of melatonin pigment within the muscle.

The individual's irises patterns are unique and structurally distinct, which remains stable throughout adult life and makes it suitable to be used for reliable automatic recognition of persons as an attractive goal. Iris recognition is employed as the most reliable and accurate biometric identification system, compared with other biometric technologies, such as speech, finger, and face recognition [1-4].

Automated iris segmentation has been an attractive topic of research in the recent past [5, 6], and many methods [7-9] have been proposed to solve the problem. The first automatic method was presented by Daugman [10] using an efficient integrodifferential operator, which is still utilized in today's most of the iris recognition systems. Image processing techniques as the first step can be applied to extract the unique pattern from the image, and encode it [11-20]. The feature extraction is another important part of Iris recognition discussed in some researches [3, 7, 10]. The shape and texture features are useful for identifying the Iris region's geometric properties. In contrast, FFT and Wavelet feature mostly represent energy distributions and convergence at different frequencies and boundaries, making them helpful in quantifying the Iris region heterogeneity [21-24]. Also, GLDM and GLCM features are useful for identifying and computing the textures for assessing the heterogeneity in Iris region textual details [25-29].

In this paper, we proposed a four-stage machine learning-based Iris recognition using eye images. In the first step, we implemented the Iris region segmentation function by applying two techniques of relative total variation and Coarse Iris Localization, improved by shearlet transform in the edge detection stage. We also used Daugman's rubber sheet model to transfer the detected Iris region to a rectangular form. In the second step, we proposed a computation and analysis scheme to generate a feature pool of spatial and frequency components from each segmented Iris region. In the third step, we reduced the feature pool size to find the optimal feature fusion and remove less important features by employing the Kernel-PCA technique. The final feature vector was fed to a multi-layer neural network, proposed in the fourth step.

## II. Materials and Methods

### A. Dataset

The proposed approach's performance was evaluated on available databases CASIA-Iris-Interval version 4 under near-infrared, including 100 people and ten eye images per person, 2000 images in total [30]. One close-up iris camera is used to capture Iris images of this dataset. The camera, used in this dataset, employs a circular NIR LED array, with suitable luminous flux for the imaging. Because of this novel design, the iris camera can capture very clear iris images (Fig.1). CASIA-Iris-Interval is well-suited for analyzing the detailed texture features of iris images.

### B. Iris Region Segmentation

The preprocessing part consisted of enhancement, noise removal, and reflection removal. Some factors, including angle and intensity of illumination source, can leave an undesirable effect on the quality of iris image and, as a result, on segmentation and recognition accuracy. To address this problem, the Single Scale Retinex (SSR) method and normalizing eye image illumination are proposed in [31].

After enhancement, we applied a median filter to the iris image to remove isolated noisy pixels(Fig. 2b). Then we removed undesired reflection, which occurs under a less-constrained imaging environment, by a thresholding process so that pixels whose gray levels are higher than the highest threshold were moderated (Fig. 2c).

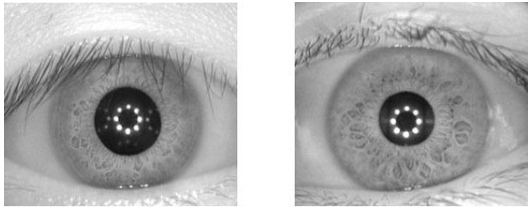

Fig. 1: An example of iris images in the CASIA-Iris-Interval dataset.

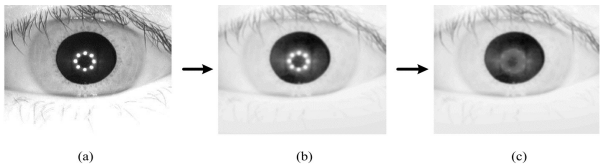

Fig. 2: Preprocessing of a single eye image. a) Original image. b) Enhanced version. c) Reflection removal.

To identify iris boundaries and separate it from other parts like the pupil, and eyelids, we used an approach proposed in [32]. This segmentation method, employing relative total variation and Coarse Iris Localization techniques, is able to segment the iris area in the target eye image effectively. However, it utilizes conventional edge detection techniques such as Canny and Sobel to find edges in the image. We replaced the edge detection function with a more accurate edge detection approach (Fig. 3) employing shearlet transforms [33].

Fig. 4 confirms that the proposed method localizes the distributed discontinuities in binary eye images more efficiently compared with Canny and Sobel methods. Also, Fig. 5 summarizes the steps of our Iris segmentation process that we applied to separate the Iris region from other parts of the eye image.

After segmentation, we transferred each angular segmented region to a rectangular mapped image with the fixed size for all cases using Daugman's rubber sheet model (Fig. 5 and Fig. 6), where radius r is between 0 and 1, and θ is 1 to 360 degree.

C. Feature extraction

In this step, we used a scheme to extract 252 features from both the spatial and frequency domains including shape&density, Gray Level Difference Method (GLDM), Gray-Level Co-Occurrence Matrix (GLCM) method, Fast Fourier Transform (FFT), and Wavelet transform (Fig. 8 and Fig. 9). For each group and each subsection, we measured 14 features applying the same statistical calculations such as Area, Mean, Std, Max, Min, Mean Deviation, Energy, Entropy, Kurtosis, Skewness, Range, RMS, Median, and Uniformity resulting in 252 features for each iris region totally. We performed GLCM and GLDM techniques in four different directions, and Wavelet transforms were also implemented in eight sub-bands.

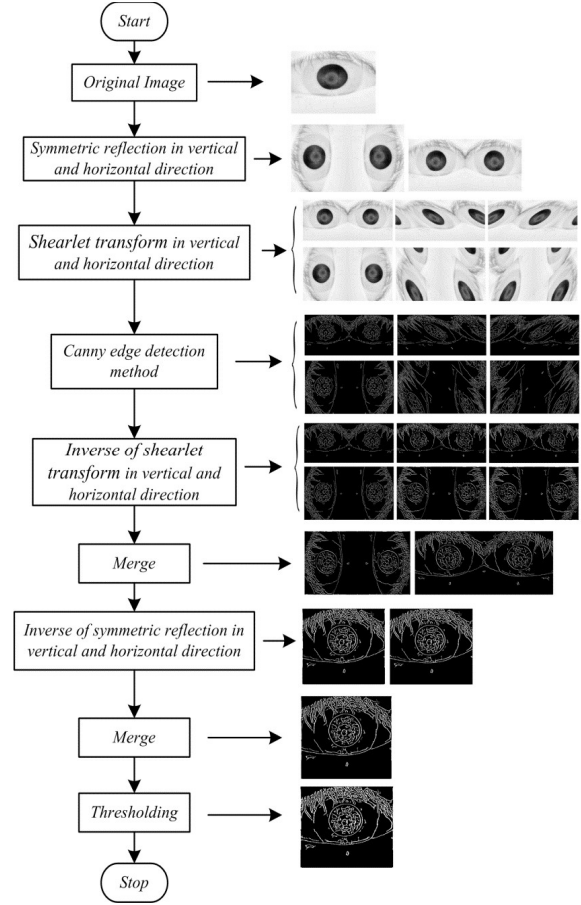

Fig. 3: The edge detection method based on shearlet transforms.

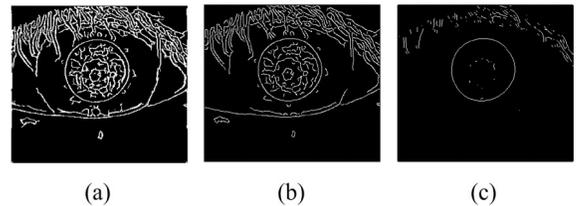

Fig. 4: Edge detection results. a) Proposed method. b) Canny method. c) Sobel method

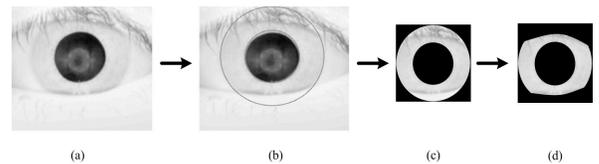

Fig. 5: Iris segmentation process. a) Preprocessed image. b) Iris and pupil circle localization. c) Extracted iris area d) Eyelids removal.

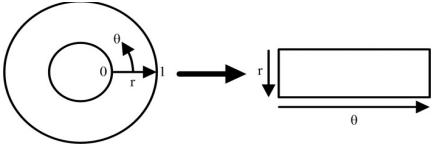

Fig. 6: Daugman's rubber sheet model.

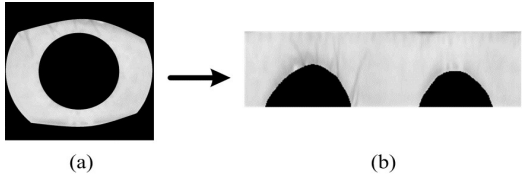

(a)  (b)

Fig. 7: Spatial transformation. a) Segmented iris image. b) Transformed rectangular image.

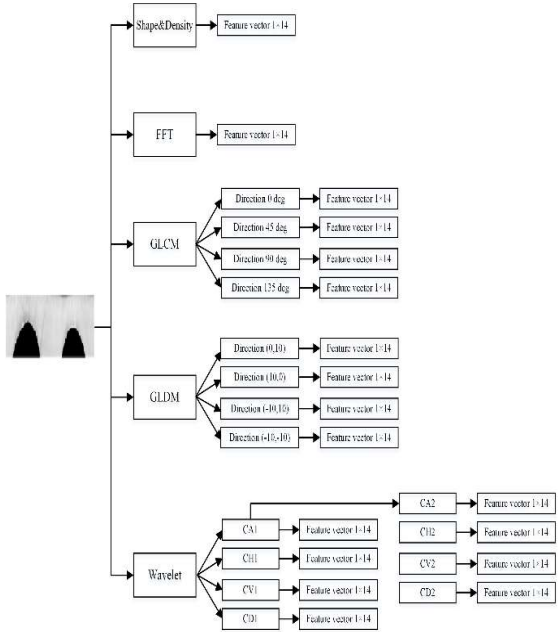

Fig. 8: The proposed feature extraction scheme

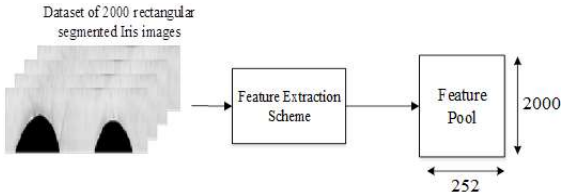

Fig. 9: Creating feature pool by feeding 2000 rectangular segmented Iris images to the proposed feature extraction model

### D. Feature reduction

Given the initial pool of 252 features, we used the Kernel-PCA technique to discard non-useful features and find an optimal feature fusion. By engaging the PCA technique performed in a kernel Hilbert space, and then select the most important features [34], the original feature pool was converted to 100 new synthetic features (Fig. 10). We finally applied this new optimal feature vector in the classification and recognition step.

### E. Machine learning-based classifier

For the classification and recognition task, we implemented a multi-layer neural network (Fig. 10) using Keras library in python. Nowadays, neural networks and deep learning models are important parts of detection, prediction, classification, and recognition systems with different applications [35-41]. Our designed model was built by applying two hidden layers (with 1000 and 400 neurons, respectively) followed by one output classifier with 200 output class labels matching to 200 people in our dataset. Also, we used drop-out techniques to decrease the risk of overfitting during the training process. The total number of pf parameters in this neural network model is 581,600, which is effectively lower than deep learning models usually used in classification and recognition tasks such as AlexNet, Vgg, GoogleNet, and ResNet.

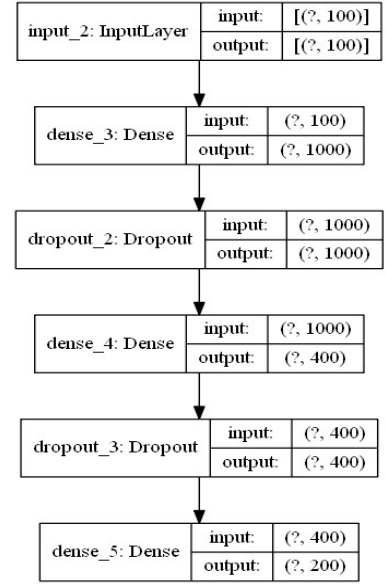

Fig. 10: The proposed multi-layer neural network structure used for the recognition task

## III. RESULTS

### A. Analysis of extracted features

Fig. 11a shows the Pearson correlation coefficient matrix of the original extracted 252 features for 2000 observations. This map reveals that the Wavelet group has less dependence compared to other groups. Also, the histogram of correlation coefficients is shown in Fig. 11b indicates that the feature pool generated in our study can provide a comprehensive representation of the iris region because more than 80% of the correlation coefficient values were less than 0.5.

We used the AUC value (Area under the ROC curve) as an indicator to compare the discrimination power of different single (e.g., RMS, mean_wavelet, std_GLDM) and group (e.g.,

Shape&Density, FFT, GLCM) features shown in Fig. 12. We sorted all the features in the order of their average AUC value (Fig. 12a). As seen, most of the features (148 out of 252) showed the AUC value of more than 0.6, while Entropy, Mean_FFT, and Energy_Wavelet are the top three features with an AUC value of 0.88±0.06, 0.86±0.07, 0.84±0.08, respectively. Also, the FFT group recorded the best performance among the other features groups confirming the importance of frequency features compared to spatial features. In contrast, the lowest performance belonged to the GLCM group (Fig. 12b).

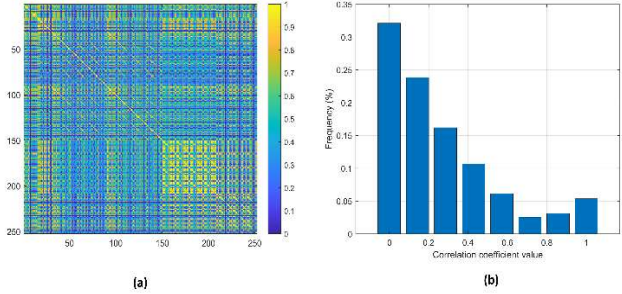

Fig. 11: Correlation analysis of features. a) The heat map view of the Pearson correlation coefficients (1-15 for Shape&Density, 16-30 for FFT, 31-90 for GLCM, 91-150 for GLDM, and 151-270 for Wavelet). b) Histogram graph of correlation coefficients.

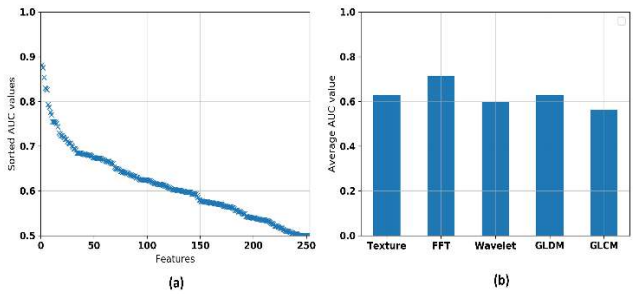

Fig. 12: The comparison of the discrimination power of the extracted features. a) AUC values of single features sorted. b) Average AUC values of different groups.

### B. Performance of Kernel-PCA features and neural network classifier

To train our proposed neural network classifier, we employed Adam optimizer to minimize categorical cross-entropy loss function during the training process. The other hyperparameters we set included MaxEpochs=100, BatchSize=8, LearningRate=0.0001, DropoutValue=0.2, ValRatio=0.2, TrainRatio=0.6, and TestRatio=0.2. Fig. 13 shows the performance of the training process comparing validation and training loss function values converged at 35 epochs with a score value of 0.24 and the accuracy of 97.6% for the test set.

Table 1 shows the average performance of our proposed iris recognition scheme confirming that the synthetic optimal feature group (K-PCA features) achieves a considerably better performance than the best single feature in Fig. 12a.

TABLE I. EVALUATION OF AVERAGE METRICS OF 400 TEST SAMPLES

| *Precision* | *Sensitivity* | *F-score* | *Support* |
|---|---|---|---|
| 0.97 | 0.95 | 0.96 | 400 |

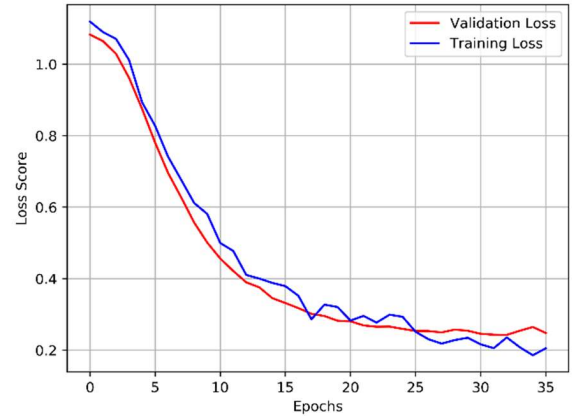

Fig. 13: Loss score graph of the training process for designed machine learning classifier

### IV. CONCLUSION AND DISCUSSION

In this study, we showed that our proposed machine learning scheme of iris region segmentation, feature extraction, and neural network classifier could accurately recognize and classify Iris images. Unlike various previously proposed machine learning schemes that use the texture-based features in the spatial domain, we computed image features in both the spatial and frequency domains. Also, since creating optimal and most effective image features is one of the most critical tasks in building machine learning-based classifiers, we investigated the importance of utilizing a data reduction method to select more correlated and optimal features. The results proved that our feature reduction method decreases the size of feature space and can replace the new smaller feature vector with more correlated information and a lower redundancy.

### REFERENCES

1. P. Polash, M. Monwar et al., "Human iris recognition for biometric identification," in Computer and information technology, ICCIT 2007. 10th international conference on. IEEE, pp. 1–5, 2008.
2. J. Daugman, "How Iris Recognition Works," IEEE Transaction on Circuits and Systems for Video Technology, Vol 14, No. 1, pp 21-30, 2004.
3. A. K. Mobarakeh et al., "A Comparative Study of Different Dimension Reduction Techniques for Face Recognition," American Journal of Computer Science and Engineering, Vol. 4, No. 1, pp. 1-7, 2017.
4. H. Procenca, "Iris Recognition: On the Segmentation of Degraded Images Acquired in the Visible Wavelength," IEEE Transactions on Pattern Analysis and Machine Intelligence, Vol. 32, No. 8, pp 1502-1516, 2010.
5. A. Kumar, T.-S. Chan, "Iris recognition using quaternionic sparse orientation code (QSOC)," Proc. CVPR 2012, pp. 59-64, CVPRW 2012, Providence, 2012.

6. A. Kumar, T.-S. Chan, C. W. Tan, "Human identification from at-a-distance face images using sparse representation of local iris features," Proc. ICB, 2012, pp. 303-309, 2012.
7. N. Bhatia and M. Chhabra, "Improved Hough Transform for Fast Iris Detection," Second IEEE International Conference on Signal Processing Systems, Vol. 5, pp. 172 -176, 2010.
8. Md. R. Islam et al., "Partial Iris Image Recognition Using Wavelet Based Texture Features," International Conference on Intelligent and Advanced Systems (ICIAS), 2010.
9. C. W. Tan and A. Kumar, "Unified framework for automated iris segmentation using distantly acquired face images," IEEE Trans. Image Process., Vol. 21, No. 9, pp. 4068-4079, 2012.
10. J. Daugman, "How iris recognition works," Circuits and Systems for Video Technology, IEEE Transactions on, Vol. 14, No. 1, pp. 21–30, 2004.
11. M. Heidari et al., "Prediction of breast cancer risk using a machine learning approach embedded with a locality preserving projection algorithm," Physics in Medicine & Biology Vol. 63, No. 3, 2018.
12. Y. Du et al., "Classification of tumor epithelium and stroma by exploiting image features learned by deep convolutional neural networks," Ann. Biomed. Eng., Vol. 46, No. 12, pp. 1988–1999, 2018.
13. Y. Du et al., "A performance comparison of low-and high-level features learned by deep convolutional neural networks in epithelium and stroma classification," SPIE Medical Imaging 2018: Digital Pathology, International Society for Optics and Photonics, Vol. 10581, 2018.
14. X. Chen et al., "Applying a new quantitative image analysis scheme based on global mammographic features to assist diagnosis of breast cancer," Computer methods and programs in biomedicine, Vol. 179, 2019.
15. A. Zargari et al., "Applying a new unequally weighted feature fusion method to improve CAD performance of classifying breast lesions," SPIE Medical Imaging 2018, International Society for Optics and Photonics, Vol. 10575, 2018.
16. S. M. Khaniabadi et al., "The Performance Assessment of Curve Fitting Tools of Underwater Object with Glass Condition Using Stereo Vision," International Journal of Computer Science and Control Engineering, Vol. 4, No. 3, pp. 14-17, 2016.
17. M. Heidari et al., "Assessment of a quantitative mammographic imaging marker for breast cancer risk prediction," SPIE Medical Imaging 2019, International Society for Optics and Photonics, Vol. 10952, 2019.
18. A. Zargari et al., "Assessing the performance of quantitative image features on early stage prediction of treatment effectiveness for ovary cancer patients: a preliminary investigation," SPIE Biophotonics and Immune Responses XIII, International Society for Optics and Photonics, Vol. 10495, 2018.
19. S. M. Khaniabadi et al., "The Performance Evaluation of Two Different Distance Estimation Tools Under Unclean Water Using Stereo Vision," International Journal of Computer Science and Control Engineering, Vol. 5, No. 1, pp. 1-4, 2017.
20. M. Heidari et. al, "Improving performance of breast cancer risk prediction using a new CAD-based region segmentation scheme," SPIE Medical Imaging, International Society for Optics and Photonics, Vol. 10575, 2018.
21. Vanaja Roselin.E.C, Dr.L.M.Waghmare, "Pupil detection and feature extraction algorithm for Iris recognition", AMO-Advanced Modeling and Optimization, Vol. 15, No. 2, 2013.
22. A. Zargari et al., "Prediction of chemotherapy response in ovarian cancer patients using a new clustered quantitative image marker," Physics in Medicine and Biology, Vol. 63, No. 15, 2018.
23. M. Heidari, S. Ghaemmaghami, "Universal image steganalysis using singular values of DCT coefficients," 10th International ISC Conference on Information Security and Cryptology, 2013.
24. H. Pourghassem, and H. Ghassemian, "Content-based medical image classification using a new hierarchical merging scheme," Computerized Medical Imaging and Graphics, Vol. 32, No. 8, pp. 651-661, 2008.
25. R. N. Khushaba et al., "Novel feature extraction method based on fuzzy entropy and wavelet packet transform for myoelectric Control," in International Symposium on Communications and Information Technologies. 2007.
26. A. Z. Khuzani et al., "Covid-classifier: An automated machine learning model to assist in the diagnosis of covid19 infection in chest x-ray images," medRxiv, Cold Spring Harbor Laboratory Preprints, 2020.
27. A. Zargari et al., "Fire detection in video sequences using a machine learning system and a clustered quantitative image marker," IEEE global humanitarian technology conference (GHTC), 2019.
28. J. K. Kook, and P. H. Wook, "Statistical textural features for detection of microcalcifications in digitized mammograms," IEEE Transactions on Medical Imaging, Vol. 18, No. 3, pp. 231-238, 1999.
29. M. M. Saleck, A. ElMoutaouakkil, M. Mouçouf, "Tumor Detection in Mammography Images Using Fuzzy C-means and GLCM Texture Features," 14th International Conference on Computer Graphics, Imaging and Visualization, pp. 122 – 125, 2017.
30. Biometrics Ideal Test, CASIA-Iris-Interval version 4: http://www.idealtest.org/dbDetailForUser.do?id=4.
31. D. H. Brainard and B. A. Wandell, "Analysis of the retinex theory of color vision," J. Optical Soc. Am. A., Vol. 3, no. 10, pp. 1651-1661, 1986.
32. Zijing Zhao, Ajay Kumar, "An Accurate Iris Segmentation Framework under Relaxed Imaging Constraints using Total Variation Model," IEEE International Conference on Computer Vision (ICCV), pp. 3828 – 3836, 2015.
33. Wang.-Q. Lim, "The discrete shearlet transform: a new directional transform and compactly supported shearlet frames," IEEE Trans. Image Process. Vol. 19, No. 5, pp. 1166–1180, 2010.
34. C. Liu, T. Zhang, D. Ding, C. Lv, "Design and application of Compound Kernel-PCA algorithm in face recognition," 35th Chinese Control Conference, pp. 4122 – 4126, 2016.
35. M. Heidari et al., "Applying a machine learning model using a locally preserving projection based feature regeneration algorithm to predict breast cancer risk", SPIE Medical Imaging, International Society for Optics and Photonics, Vol. 10579, 2018.
36. W. Liu et al., "Utilizing deep learning technology to develop a novel CT image marker for categorizing cervical cancer patients at early stage," SPIE Biophotonics and Immune Responses XIV, International Society for Optics and Photonics, Vol. 10879, 2019.
37. M. Heidari et al., "Improving performance of CNN to predict likelihood of COVID-19 using chest X-ray images with preprocessing algorithms," arXiv preprint arXiv:2006.12229, 2020.
38. L. Jia et al., "A rule-based method for automated surrogate model selection," Advanced Engineering Informatics, Vol. 45, 2020.

39. R. Alizadeh et al. "Ensemble of surrogates and cross-validation for rapid and accurate predictions using small data sets," AI EDAM, Vol. 33, No. 4, pp. 484-501, 2019.
40. R. Alizadeh et al., "Managing computational complexity using surrogate models: a critical review," Research in Engineering Design, Vol. 31, No. 3, pp. 275-298, 2020.
41. H. Z. Sabzi et al. "Integration of time series forecasting in a dynamic decision support system for multiple reservoir management to conserve water sources," Energy Sources, Part A: Recovery, Utilization, and Environmental Effects, Vol. 40, No. 11, pp. 1398-1416, 2018.